\documentclass[a4paper,fleqn]{cas-dc}

\usepackage[authoryear,longnamesfirst]{natbib}

\usepackage{algorithmic}
\usepackage{algorithm}
\usepackage{subcaption}
\usepackage{graphicx}
\usepackage{textcomp}
\usepackage{xcolor}
\usepackage{hyperref}
\usepackage{physics}
\usepackage{booktabs}

\def\tsc#1{\csdef{#1}{\textsc{\lowercase{#1}}\xspace}}
\tsc{WGM}
\tsc{QE}

\begin{document}
\let\WriteBookmarks\relax
\def\floatpagepagefraction{1}
\def\textpagefraction{.001}
\let\printorcid\relax 

\shorttitle{View-Aware Quantization.}    

\shortauthors{K. Ardif et~al.}  

\title [mode = title]{Communication-Efficient Federated Multi-View
Clustering via View-Aware Quantization.}

\author[1]{Khadija Ardif}\cormark[1]
\ead{ardif.khadija@doctorant.inpt.ac.ma}
\credit{Conceptualization, Methodology, Software, Investigation, Formal analysis, Validation, Visualization, Writing – original draft, Writing – review \& editing}
\author[1]{Meryeme Ayache}
\ead{ayache@inpt.ac.ma}
\credit{Supervision, Conceptualization, Writing - review \& editing, Validation}
\author[2]{Nidal Nasser}
\ead{nnidal@gmail.com}
\credit{Supervision, Writing - review \& editing, Validation}
\author[3]{Ouiame Marnissi}
\ead{ouiame.marnissi@um6p.ma}
\credit{Conceptualization, Writing - review \& editing, Validation}

\affiliation[1]{organization={STRS Lab, National Institute of Posts and Telecommunications INPT}, 
                city={Rabat},
                country={Morocco}}
\affiliation[2]{organization={College of Engineering \& Advanced Computing, Alfaisal University}, 
                city={Riyadh},
                country={Saudi Arabia}}
\affiliation[3]{organization={College of Computing Mohammed VI, Polytechnic University}, 
                city={Ben Guerir},
                country={Morocco}}                

\cortext[1]{Corresponding author}

\begin{abstract}
Federated multi-view clustering (FedMVC) has recently attracted increasing attention 
for its ability to exploit complementary clustering structures across multiple clients 
while preserving data privacy. However, existing FedMVC methods often suffer from 
high communication overhead due to the repeated transmission of high-dimensional 
representations, which limits their scalability in practical federated systems. 
Moreover, different views may contribute unequally to the clustering performance, 
calling for adaptive communication strategies.

In this paper, we propose a view-aware quantization algorithm for 
communication-efficient FedMVC. By dynamically 
assigning quantization bits based on iteratively updated view importance scores, 
the proposed algorithm effectively balances clustering performance and communication 
efficiency under a global bit budget constraint. To validate its generality, 
we integrate our approach into four state-of-the-art FedMVC 
methods. Extensive experiments on multiple benchmark datasets demonstrate that 
view-aware quantization achieves up to 6.59× communication reduction while 
maintaining and in some cases surpassing baseline clustering performance, 
confirming the effectiveness and scalability of the proposed approach.
\end{abstract}



\begin{NoHyper}
    \maketitle
\end{NoHyper}

\section{INTRODUCTION}

Federated multi-view clustering (FedMVC) provides an effective framework for collaborative 
clustering under privacy constraints \cite{huang2022efficient}, but its practical deployment is often hindered 
by high communication costs. In this work, we focus on improving the 
communication efficiency of FedMVC by proposing a view-aware 
quantization strategy that adaptively allocates communication precision according 
to the heterogeneous contribution of each view.

Multi-view clustering exploits complementary information from multiple feature 
representations to uncover robust clustering structures \cite{chen2022representation}. In real-world applications 
such as multimedia analysis, healthcare, and recommendation systems, data are 
naturally represented by heterogeneous views. Traditional methods, however, 
assume centralized access to all views, which is often impractical due to privacy 
and regulatory constraints. Federated learning (FL) addresses these issues by 
enabling collaborative training without sharing raw data. Building on this \cite{mcmahan2017communication}, 
FedMVC has gained increasing attention for mining 
shared clustering structures across distributed clients while preserving data privacy \cite{chen2023federated}. 
Despite its advantages, FedMVC introduces significant communication overhead, 
which plays a critical role in determining its scalability and practicality.

Communication overhead is a major bottleneck in FL systems, 
particularly for clustering tasks that require repeated exchange of high-dimensional 
latent representations and cluster centroids across multiple communication rounds. 
In FedMVC, this challenge is amplified by the presence of 
multiple views, each transmitting view-specific representations to the server. 
Moreover, views are often heterogeneous in terms of data quality, feature 
dimensionality, and contribution to clustering performance. Treating all views 
equally during communication therefore leads to inefficient bandwidth usage, where 
less informative views consume the same resources as more important ones. These 
challenges call for communication-efficient FedMVC methods that account for view 
heterogeneity when allocating transmission resources.

To alleviate communication costs in FL, various compression 
techniques such as quantization and sparsification have been widely studied ~\cite{alistarh2017qsgd, reisizadeh2020fedpaq, wangni2018gradient}. 
While existing compression methods have been proposed for gradient-based federated 
optimization, they are not designed for the specific requirements of FedMVC. When applied to FedMVC, these approaches suffer from several 
limitations. First, they do not account for the heterogeneous contribution of 
each view, leading to suboptimal allocation of communication resources. Second, 
uniform compression applies the same bit budget to all views regardless of their 
importance, which may disproportionately degrade the quality of more informative 
representations. Third, sparsification techniques are primarily designed for 
gradient transmission and do not naturally extend to the latent representations 
exchanged in clustering-based federated systems, as zeroing out specific dimensions 
may destroy the underlying clustering structure. As a result, a clear gap remains 
between existing communication-efficient FL techniques and the 
specific requirements of FedMVC, calling for adaptive compression strategies that 
explicitly account for view importance.\\
To address the above challenges, this paper proposes 
a View-Aware Quantization (VAQ) algorithm for 
communication-efficient FedMVC. Rather than uniformly 
applying quantization across all views, VAQ introduces 
a dynamic view importance estimation mechanism combined 
with a principled bit allocation strategy, formulated 
as a constrained optimization problem, to adaptively 
assign communication precision according to each 
view's contribution to the global clustering objective.
The main contributions of this work are summarized as follows:
\begin{itemize}
    \item We formalize the problem of view-aware communication efficiency in FedMVC as a constrained optimization problem, where quantization bits are allocated across views under a global bit budget constraint, and define a communication overhead metric to quantify transmission costs across multiple communication rounds.
    \item We propose a VAQ algorithm that dynamically allocates quantization bits across views based on iteratively updated importance scores, effectively balancing clustering performance and communication efficiency.
    \item We validate the generality of the proposed approach by integrating it into four state-of-the-art FedMVC methods. Extensive experiments on five benchmark datasets demonstrate up to 6.59× communication reduction while maintaining and in some cases surpassing baseline clustering performance.
    \item We conduct an ablation study to validate the 
    contribution of each component of the proposed VAQ 
    algorithm, including the view-aware bit allocation 
    strategy and the dynamic importance score update. 
    Sensitivity analysis further demonstrates the 
    robustness of the proposed approach with respect 
    to the weighting factor $\alpha$.
\end{itemize}
The rest of this paper is organized as follows. Section ~\ref{RELATED_WORK} reviews related work on 
FedMVC and communication-efficient FL. 
Section ~\ref{sec:problem_formulation} presents the problem formulation, including the communication overhead 
metric and the optimization problem. Section ~\ref{sec:Proposed} describes the proposed VAQ algorithm. Section ~\ref{sec:results} describes the experimental setup, datasets, 
and evaluation metrics and discusses the experimental results. Finally, 
Section ~\ref{sec:conclusion} concludes the paper and outlines directions for future work.

\section{RELATED WORK}
\label{RELATED_WORK}
\subsection{Communication-Efficient Federated Learning}
Communication overhead remains one of the main bottlenecks in FL, particularly 
in large-scale and heterogeneous environments. Several studies have 
proposed compression strategies to reduce transmission costs while maintaining model 
performance.\\ 
Quantization-based approaches reduce communication by encoding model 
updates using low-bit representations. Authors in ~\cite{alistarh2017qsgd} 
proposed QSGD, a family of gradient compression schemes for parallel SGD that 
provides convergence guarantees for both convex and non-convex objectives 
while allowing users to smoothly trade off communication bandwidth against 
convergence time. Reisizadeh the work in ~\cite{reisizadeh2020fedpaq} proposed FedPAQ, 
a communication-efficient FL method that combines periodic 
averaging, partial device participation, and quantized message-passing to reduce 
the communication overhead between edge devices and the parameter server.\\
Complementary to quantization, sparsification-based approaches reduce 
communication by transmitting only the most significant 
gradient components. In this direction, the method 
in \cite{aji2017sparse} introduced 
accelerating distributed SGD by exchanging sparse updates instead of dense ones, 
mapping the 99\% smallest gradient updates (by absolute value) to zero to reduce 
communication, a technique that can be combined with quantization for further 
compression. The study in ~\cite{stich2018sparsified} demonstrated that SGD with 
k-sparsification or compression converges at the same rate as vanilla SGD when 
equipped with error compensation, effectively reducing communication by a factor of 
the problem dimension without sacrificing convergence speed.\\ 
Building on these two families of methods, more 
recent works have proposed adaptive compression strategies that dynamically combine 
or adjust the compression rate based on training dynamics. 
These methods adapt quantization levels or sparsification ratios according 
to gradient magnitude, client heterogeneity, or communication constraints, 
achieving better communication-performance trade-offs \cite{konevcny2016federated,sattler2019robust}.
For instance, the approach in ~\cite{marnissi2024adaptive} formulated 
an adaptive sparsification and quantization scheme 
that jointly optimizes both compression mechanisms 
to enhance energy efficiency in FL. 
Despite their effectiveness in supervised FL, these methods are designed for 
gradient transmission and assume homogeneous 
feature representations across clients. As a result, they do not account for 
the heterogeneous contribution of different views in a multi-view clustering 
setting, leaving a clear gap that motivates the present work.

\subsection{Communication Efficiency in Federated Multi-View Clustering}
Despite the growing interest in communication-efficient 
FL, very few works have explicitly 
addressed this challenge in the context of FedMVC. 
Most existing FedMVC methods focus on clustering 
performance and privacy preservation without explicitly 
considering communication overhead.
Recent works have approached the communication efficiency problem in FedMVC from 
various directions. the authors in~\cite{liu2025communication} recently proposed a 
communication-efficient FedMVC framework that reduces transmission costs by 
limiting the exchange of data representations across communication rounds.
Similarly, the method in~\cite{chen2025effective} developed ESFMC, which 
leverages an information-theoretic perspective to split 
locally extracted features, retaining sensitive information 
locally and sharing only task-relevant features with the 
server. While this selective sharing implicitly reduces 
the volume of transmitted data, its primary objective 
is to mitigate privacy risks rather than to explicitly 
optimize communication efficiency.
More recently, EFIMVC~\cite{liu2025efficient} introduced a localized optimization 
strategy that employs view-specific and shared anchor graphs as communication 
variables, significantly reducing communication costs 
while ensuring theoretical convergence and enhancing 
privacy by avoiding the transmission of sensitive 
embeddings. EFIMVC also extends to scenarios with 
missing views, making it a practical solution for 
real-world applications. 
In a similar vein, the work in~\cite{feng2025federated} introduced FMVC-IAI, 
which constructs a consensus graph structure through 
complementary multi-view data and shares anchor graphs 
instead of full adjacency graphs as communication 
variables, reducing both communication costs and 
privacy leakage risks. While this approach effectively 
reduces the volume of transmitted graph data, it does 
not address quantization of latent representations 
exchanged during the clustering process.\\
In summary, while existing FedMVC methods have made 
notable progress in reducing communication overhead 
through selective feature sharing, anchor graph-based 
communication, and localized optimization strategies, 
none of them explicitly addresses quantization as a 
compression mechanism. Moreover, no existing work 
accounts for the heterogeneous contribution of 
different views when allocating communication 
resources. This motivates the proposed view-aware 
quantization algorithm, which adaptively assigns 
quantization bits to each view according to its 
dynamically estimated importance, filling a clear 
gap in the existing literature on communication-efficient FedMVC.
\section{PROBLEM FORMULATION}
\label{sec:problem_formulation}

This section presents the formal problem formulation. We begin by introducing 
the system model and notations, and then define the FedMVC 
objective. Next, we describe the communication and compression models, along with 
the communication overhead metric. Finally, we present the resulting optimization 
problem. The main notations are summarized in Table \ref{tab:notation}.
In this work, we consider a federated setting in which each client owns a single 
view of the same set of samples, while observing heterogeneous feature spaces.
\begin{table}[htbp]
\caption{Summary of Main Notations}
\label{tab:notation}
\centering
\resizebox{\columnwidth}{!}{%
\begin{tabular}{|c|c|}
\hline
\textbf{Notation} & \textbf{Definition} \\
\hline
V & Number of federated clients (views)\\
\hline
v & Client / view index\\
\hline
n & Number of shared samples\\
\hline
T & Total number of federated communication rounds\\
\hline
K & number of clusters\\
\hline
t & communication round index\\
\hline
$d_v$ & Feature dimension of the local representation at client v\\
\hline
$\alpha \in [0,1] $ & Trade-off parameter balancing compactness and centroid stability in the dynamic importance score\\
\hline
$\epsilon$ & Small positive constant for numerical stability\\
\hline
$X^v \in \mathbb{R}^{n \times d_v}$ & Local data at client v\\
\hline
$Z^v \in \mathbb{R}^{n \times r}$ & Latent representation of client v\\
\hline
$C_v^{(t)} \in \mathbb{R}^{K \times r}$ & Local centroid matrix of client v at round t\\
\hline
$Y_v^{(t)} \in \{0,1\}^{n \times K}$ & Local assignement matrix at round t\\
\hline
$Y^{(t)} $ & Global assignement matrix at round t\\
\hline
$\norm{.}_F$ & Frobenius norm\\
\hline
B & Total communication bit budget per round\\
\hline
$b_v^{(t)}$ & number of bits allocated to client v at round t\\
\hline
$Q_b(.)$ & quantization operator using b bits\\
\hline
$S_v^{(t)}$ & Dynamic importance score at round t\\
\hline
$I_v^{\mathrm{init}}$ & Initial importance score computed during initialization\\
\hline
$\mathcal{L}_{compact,v}^{(t)}$ & Compactness metric\\
\hline
$S_{drift,v}^{(t)}$ & Centroid stability metric\\
\hline
$\tilde{C}_v^{(t)}$ & Quantized centroids\\
\hline
CO & Communication overhead\\
\hline
\end{tabular}%
}
\end{table}

\subsection{Learning Model}
Let $\{X^1,X^2,\dots,X^V\}$ denote multi-view data distributed over V isolated clients 
$\{C_1,C_2,\dots,C_V\}$, where client $C_v$ exclusively holds view $X^v$. A trusted server 
S is responsible for aggregating client-side information to derive a global 
clustering solution, while raw data remain decentralized.
All clients share the same samples but observe different feature spaces. 
Formally, client $v \in \{1,\dots,V\}$ owns a local dataset $X^v \in \mathbb{R}^{n \times d_v}$, 
where $d_v$ denotes the feature dimension of view v. Raw data never leave the clients.
Each client trains a local encoder 
$f_v\!(\cdot)$ and produces a latent representation $Z_v = f^v(X^v) \in \mathbb{R}^{n \times r}$
where r is the shared latent dimension.
In addition, each client maintains a local centroid matrix $C_v \in \mathbb{R}^{K \times r}$
with K clusters.
The goal is to learn a consensus clustering structure across views without sharing 
raw data, while minimizing communication costs.\\

Inspired by \cite{liu2025communication}, we formulate the 
local clustering objective in FedMVC as : 
Each client $v$ holds a local dataset $X^v \in \mathbb{R}^{n 
\times d_v}$, which is first encoded into a latent representation 
$Z^v \in \mathbb{R}^{n \times r}$ through a local encoder. The 
local clustering objective at client $v$ aims to find the optimal 
centroid matrix $C_v \in \mathbb{R}^{K \times r}$ and assignment 
matrix $Y_v \in \{0,1\}^{n \times K}$ that minimize the 
reconstruction error between the latent representations and their 
assigned cluster centroids:

\begin{equation}
    \min_{C_v, Y_v} \|Z^v - Y_v C_v\|_F^2
    \label{eq:local_clustering}
\end{equation}

At each communication round $t$, the local centroid matrix is 
updated by solving:

\begin{equation}
    C_v^{(t)} = \arg\min_{C} \|Z^v - Y_v^{(t-1)} C\|_F^2
    \label{eq:centroid_update}
\end{equation}

where $Y_v^{(t-1)} \in \{0,1\}^{n \times K}$ denotes the local 
assignment matrix from the previous round. This update corresponds 
to a least-squares solution that repositions each centroid to the 
mean of its currently assigned samples.

\subsection{Communication Model}
\begin{flushleft}
During federated training, clients and the central server communicate periodically over multiple communication rounds. At each round, information is exchanged in both directions to enable collaborative clustering without sharing raw data.
\end{flushleft}
\textbf{Client-to-Server Communication.}
At each communication round $t$, the information transmitted 
by client $v$ to the server depends on the underlying FedMVC 
method. In general, two transmission schemes can be distinguished:

\begin{itemize}
    \item \textbf{Full representation exchange:} some methods 
    require each client to transmit its latent representation 
    matrix $Z^v \in \mathbb{R}^{n \times r}$ at every round, 
    along with its local centroid matrix $C_v^{(t)} \in 
    \mathbb{R}^{K \times r}$ and assignment matrix $Y_v^{(t)} 
    \in \{0,1\}^{n \times K}$.

    \item \textbf{Compact exchange:} other methods, such as \cite{liu2025communication}, transmit the full representation $Z^v$ only 
    at the first round (or during a pretraining phase), and 
    subsequently share only the centroid matrix $C_v^{(t)}$ 
    and the assignment matrix $Y_v^{(t)}$ in the following 
    rounds, significantly reducing communication overhead.
\end{itemize}

In both cases, the communication cost scales linearly with 
the number of transmitted elements and the number of bits 
used to represent each element. Formally, the per-round 
communication cost for client $v$ is:

\begin{equation}
    \text{CO}_v^{(t)} = b \cdot \left( \beta \cdot |Z^v| 
    + |C_v^{(t)}| + |Y_v^{(t)}| \right)
    \label{eq:comm_cost}
\end{equation}

\noindent where $|Z^v| = n \cdot r$, $|C_v^{(t)}| = K \cdot r$, 
$|Y_v^{(t)}| = n \cdot K$, $b$ is the number of bits per 
element, and $\beta \in \{0, 1\}$ indicates whether the latent 
representation is transmitted at round $t$.

\textbf{Server-to-Client Communication.}
After aggregating information from all clients, the server 
broadcasts the global clustering information back to all 
clients. In the majority of existing FedMVC methods, 
including \cite{liu2025communication}, the server transmits at each 
communication round $t$ the same information to all 
clients $v \in \{1, \dots, V\}$:

\begin{itemize}
    \item the consensus centroid matrix 
    $C_{\text{global}} \in \mathbb{R}^{K \times r}$, 
    aggregated from the local centroid matrices of all 
    clients,
    \item the global assignment matrix $Y^{(t)} \in 
    \mathbb{R}^{n \times K}$, representing the consensus 
    cluster memberships shared across all views.
\end{itemize}

The server-to-client communication cost per round is 
thus:

\begin{equation}
    \text{CO}_{\text{server}}^{(t)} = b \cdot \left( 
    |C_{\text{global}}| + |Y^{(t)}| \right)
    \label{eq:comm_server}
\end{equation}

\noindent where $|C_{\text{global}}| = K \cdot r$, 
$|Y^{(t)}| = n \cdot K$, and $b$ is the number of bits 
per element. Since the same consensus information is 
broadcast to all $V$ clients, the server-to-client 
communication cost is independent of the number of views.\\
\textbf{Overall Communication Overhead.}
Since both client-to-server and server-to-client 
communications occur at every round, the total 
communication overhead accumulates linearly with 
the number of communication rounds $T$. Formally, 
the total communication cost over all rounds is:

\begin{equation}
    \text{CO}_{\text{total}} = \sum_{t=1}^{T} 
    \left( \sum_{v=1}^{V} \text{CO}_v^{(t)} + 
    \text{CO}_{\text{server}}^{(t)} \right)
    \label{eq:CO_total}
\end{equation}

\noindent In particular, repeatedly transmitting 
high-dimensional latent representations and clustering 
parameters makes communication the dominant bottleneck 
in FedMVC systems, especially when the number of 
samples $n$, the representation dimension $r$, or 
the number of views $V$ is large. This observation 
motivates the use of quantization as a compression 
technique to reduce the per-element bit cost $b$, 
which directly impacts the total communication 
overhead as shown in Equations~\ref{eq:comm_cost} 
and~\ref{eq:comm_server}.
\subsection{Compression Model}

To reduce communication overhead, we apply view-adaptive quantization to the transmitted matrices.
For each view $v$:

\begin{itemize}
    \item $b_v \in [1, 16] \cap \mathbb{N}$ denotes the quantization bit-width.
    \item $Q_{b_v}(\cdot)$ denotes the quantization operator with $b_v$ bits.
\end{itemize}

Given a matrix $M_v \in \mathbb{R}^{m \times d_v}$ to be transmitted, 
its quantized version is defined as

\begin{equation}
\tilde{M}_v = Q_{b_v}(M_v).
\end{equation}

We adopt symmetric uniform scalar quantization with zero-point offset. The quantization range is defined as
\begin{equation}
    q_{\min} = -2^{b_v-1}, \quad q_{\max} = 2^{b_v-1} - 1.
    \label{eq:quant_range}
\end{equation}

Let $s_v$ denote the quantization scale (step size) and $z_v$ the zero-point, computed as
\begin{equation}
    s_v = \frac{\max(\mathbf{M}_v) - \min(\mathbf{M}_v)}{q_{\max} - q_{\min}},
    \label{eq:scale}
\end{equation}
\begin{equation}
    z_v = q_{\min} - \left\lfloor \frac{\min(\mathbf{M}_v)}{s_v} \right\rfloor.
    \label{eq:zero_point}
\end{equation}

The quantized entries are computed as
\begin{equation}
    Q_{b_v}(\mathbf{M}_v) = \text{clip}\left(\left\lfloor \frac{\mathbf{M}_v}{s_v} + z_v \right\rceil, q_{\min}, q_{\max}\right),
    \label{eq:quantization}
\end{equation}
where $\lfloor \cdot \rceil$ denotes rounding to the nearest integer, and $\text{clip}(\cdot, a, b)$ restricts values to the range $[a, b]$. Dequantization recovers the approximate values:
\begin{equation}
    \tilde{\mathbf{M}}_v = \left(Q_{b_v}(\mathbf{M}_v) - z_v\right) \cdot s_v.
    \label{eq:dequantization}
\end{equation}

The quantization error is defined as
\begin{equation}
    \mathbf{E}_v = \tilde{\mathbf{M}}_v - \mathbf{M}_v.
    \label{eq:quant_error}
\end{equation}

Under uniform quantization, the error magnitude is bounded as
\begin{equation}
    \|\mathbf{E}_v\|_\infty \leq \frac{s_v}{2}.
    \label{eq:error_bound}
\end{equation}

The allocation of $b_v$ under a global communication budget constraint will 
be addressed in Section~4.

\subsection{Optimization Problem}
The objective of this work is to achieve communication-efficient 
FedMVC while preserving clustering 
performance. At each communication round $t$, each client $v$ 
contributes to the global clustering process through the 
transmission of compressed representations. Since different 
views provide heterogeneous and complementary information, 
allocating identical communication resources to all views may 
lead to inefficient bandwidth utilization. To address this, 
we introduce a VAQ strategy that assigns 
a quantization bit-width $b_v^{(t)}$ to each view according 
to its relative importance score $S_v^{(t)}$.

Formally, the bit allocation problem is formulated as:
\begin{equation}
    \max_{\{b_v^{(t)}\}} \sum_{v=1}^{V} S_v^{(t)} \cdot b_v^{(t)} 
    \quad \text{s.t.} \quad 
    \sum_{v=1}^{V} b_v^{(t)} \leq B, \quad 
    b_v^{(t)} \in \{1, \dots, 16\},
    \label{eq:optimization}
\end{equation}
where $B$ is the total communication budget in bits, 
$S_v^{(t)}$ is the importance score of view $v$ at round 
$t$, and $b_v^{(t)}$ is the number of quantization bits 
allocated to view $v$.
The constraints in~\eqref{eq:optimization} are motivated 
as follows. The inequality constraint 
$\sum_{v=1}^{V} b_v^{(t)} \leq B$ ensures that the total 
number of transmitted bits does not exceed the global 
communication budget $B$, regardless of the number of 
views or the distribution of importance scores. The 
integer constraint $b_v^{(t)} \in \{1, \dots, 16\}$ 
imposes a minimum precision of 1 bit per view to ensure 
that all views contribute to the global clustering 
process, while the upper bound of 16 bits avoids 
excessive precision that would yield no practical 
benefit over full-precision transmission.
Under the linear utility assumption 
— higher bit-width leads to better representation quality 
— the optimal solution to \eqref{eq:optimization} allocates 
bits proportionally to view importance:

\begin{equation}
    b_v^{(t)} = \left\lfloor B \cdot 
    \frac{S_v^{(t)}}{\sum_{j=1}^{V} S_j^{(t)}} 
    \right\rfloor, \quad \forall v \in \{1, \dots, V\},
    \label{eq:bit_allocation}
\end{equation}
Since the floor operation may result in 
$\sum_{v=1}^{V} b_v^{(t)} \leq B$, the remaining bits are 
redistributed to the views with the highest importance 
scores, ensuring exact satisfaction of the budget 
constraint in \eqref{eq:optimization}.
This strategy ensures that views with higher importance 
receive more communication bandwidth, enabling adaptive 
precision across heterogeneous views while maintaining 
a fixed total communication cost $B$.

\section{PROPOSED VIEW-AWARE QUANTIZATION.}
\label{sec:Proposed}
In this section, we present the proposed View-Aware 
Quantization (VAQ) algorithm for communication-efficient 
FedMVC. Unlike uniform quantization 
schemes that assign identical bit-widths to all views, the 
proposed algorithm dynamically estimates the contribution 
of each view based on clustering compactness and centroid 
stability, and adaptively allocates quantization bits under 
a global communication budget constraint. This design 
enables adaptive precision control across heterogeneous 
views, reducing communication overhead while preserving 
clustering performance. The complete procedure is detailed 
in Algorithm~\ref{alg:VAQ-FMC}.

\subsection{Overview}

The proposed VAQ algorithm 
operates in two phases: an initialization phase that 
establishes an initial bit allocation based on 
pretrain-derived view importance scores, and a dynamic 
phase that iteratively updates the importance scores 
and bit allocation at each communication round.

\begin{figure}[]
\centering
\includegraphics[width=\linewidth]{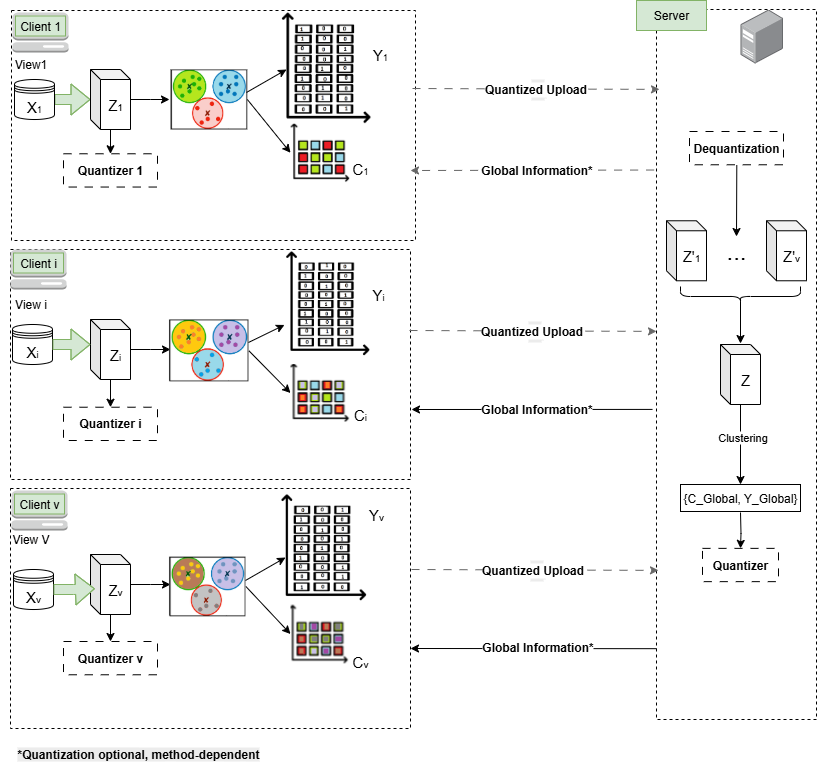}
\caption{Example of a FedMVC Architecture 
    Illustrating the Role of VAQ. Each client $i$ 
    encodes its local view $X_i$ into a latent representation $Z_i$, 
    locally clustered to yield soft assignments $Y_i$ and prototypes 
    $C_i$. Prior to transmission, a local quantizer compresses 
    $Z_i$ in a view-dependent manner. The server dequantizes the received 
    uploads to reconstruct $Z'_1, \dots, Z'_v$, aggregates them into a 
    unified representation $Z$, and performs global clustering to produce 
    $\{C_{\text{Global}}, Y_{\text{Global}}\}$, which are broadcast back 
    to refine each client's quantizer. The process iterates until 
    convergence. \textit{*Quantization is optional and method-dependent.}}
\label{fig:framework}
\end{figure}

As illustrated in Fig.~\ref{fig:framework}, 
a general FedMVC system consists of a central server 
and multiple distributed clients, where each client 
owns a single heterogeneous view of the same set of 
samples. The proposed VAQ algorithm is integrated into 
this general architecture by replacing the standard 
transmission step with a VAQ 
mechanism, as highlighted in the figure.

During the initialization phase, each client learns a 
local latent representation from its private view. To 
reduce the one-shot communication cost, each client 
computes a local view importance score reflecting 
clustering quality and representation variability. Based 
on these scores, adaptive quantization is applied before 
transmitting the local representations to the server, 
which uses them to initialize the global clustering 
variables.

After initialization, the algorithm proceeds through 
iterative communication rounds. At each round, the 
server broadcasts global clustering information to all 
clients, while each client locally updates its 
method-dependent transmission data $\mathcal{T}_v^{(t)}$ 
using its private data. The transmitted data is compressed 
using view-aware adaptive quantization based on 
dynamically updated importance scores, whereas assignment 
matrices remain uncompressed due to their discrete nature. 
By adaptively allocating communication precision according 
to view importance, the proposed algorithm effectively 
reduces communication overhead while preserving the 
complementary information across heterogeneous views.

 \subsection{Initialization Phase: View-Aware Representation Quantization}
During the initialization phase, each client transmits 
its latent representation matrix to the server to 
bootstrap the global multi-view clustering process. 
Since this transmission occurs only once but involves 
high-dimensional data, VAQ is 
applied to reduce the initial communication overhead.

To this end, each client computes an initial importance 
score reflecting the informativeness of its local view. 
Specifically, the initial importance score is designed 
to capture two complementary aspects of view quality:
\begin{itemize}
    \item \textbf{Clustering compactness:} a view is 
    more important if its local representations are 
    well-separated into distinct clusters, indicating 
    high discriminative power.
    \item \textbf{Representation variability:} a view 
    with higher variance in its latent representations 
    carries richer information and should receive more 
    communication bandwidth.
\end{itemize}
These scores are sent to the server, which performs 
adaptive bit allocation prior to the compressed 
representation transmission. Formally, the initial 
importance score of view $v$ is defined as:

\begin{equation}
\label{eq:init_importance}
I_v^{\mathrm{init}} = \lambda s_v + (1 - \lambda)\,\mathrm{Var}_v
\end{equation}
\subsection{Dynamic Importance Estimation}
In multi-view clustering, different views often provide 
complementary yet uneven contributions to clustering 
performance. Previous studies have shown that treating 
all views equally may degrade clustering quality due to 
noisy or less informative views \cite{yang2018multi,zhao2020multi}. This calls for 
an adaptive evaluation of view contribution in federated 
multi-view environments.

\begin{figure}[]
\centering
\includegraphics[width=\linewidth]{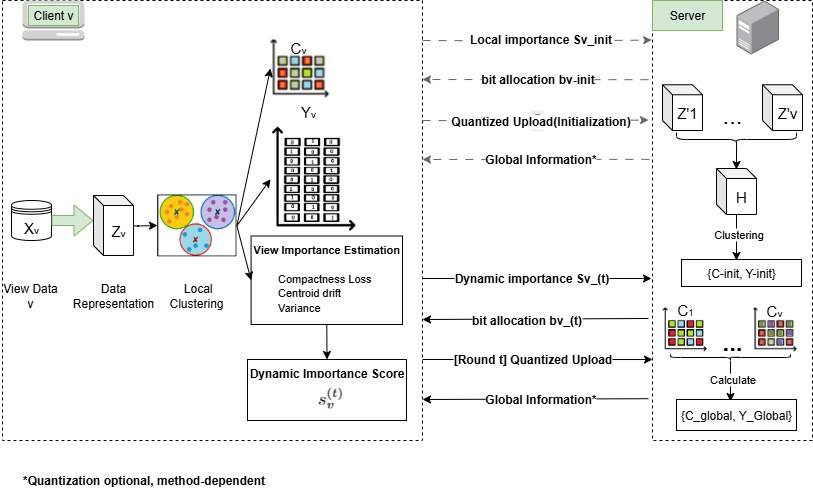}
\caption{Illustration of the proposed view-aware importance estimation
and adaptive bit allocation strategy. Each client estimates the
importance of its local view based on clustering statistics, and
the resulting score is used to dynamically determine the
quantization precision for communication-efficient transmission.}
\label{fig:Dynamic_Score}
\end{figure}

We therefore estimate, at each communication round, a 
dynamic importance score for each view based on clustering 
compactness and centroid stability. This score guides 
adaptive bit allocation, enabling communication resources 
to be preferentially assigned to more informative views. 
The overall procedure for view importance estimation and 
the resulting bit allocation strategy is illustrated in 
Fig.~\ref{fig:Dynamic_Score}.

To assess the clustering quality of each local view, we 
evaluate the compactness of its learned representations. 
Cluster compactness is a widely adopted indicator of 
clustering quality in unsupervised learning \cite{rousseeuw1987silhouettes}.
lower intra-cluster dispersion reflects stronger 
representation consistency and discriminative power. A 
view producing compact, well-separated clusters is 
therefore expected to contribute more reliably to the 
global clustering objective. Specifically, the compactness 
loss quantifies the intra-cluster dispersion between 
latent representations and their assigned centroids, 
computed as:
\begin{equation}
\label{eq:compactness}
\mathcal{L}_{compact,v}^{(t)}
=
\frac{1}{n_k}
\| Z_v - Y^{(t-1)} C_v^{(t)} \|_F^2
\end{equation}

Beyond clustering compactness, the stability of local 
updates also reflects the importance of a view. In 
federated optimization, informative clients tend to 
induce larger updates to the global model, and recent 
studies have shown that update magnitude and drift can 
effectively characterize client contribution and data 
heterogeneity \cite{karimireddy2020scaffold}.
Inspired by this observation, we 
introduce the centroid drift metric, which measures the 
variation of local centroids between consecutive 
communication rounds. A larger drift indicates that the 
corresponding view actively contributes to the global 
clustering updates, and is computed as:
\begin{equation}
\label{eq:drift}
S_{drift,v}^{(t)}
=
\| C_v^{(t)} - C_v^{(t-1)} \|_F^2
\end{equation}

To jointly capture clustering quality and training 
dynamics, we combine compactness and centroid drift into 
a unified dynamic importance score, capturing both the 
static quality and the dynamic behavior of each view. 
Specifically, the importance score of view $v$ at 
communication round $t$ is defined as a weighted 
combination of its compactness loss and centroid drift:
\begin{equation}
\label{eq:importance}
s_v^{(t)}
=
\alpha \frac{1}{\mathcal{L}_{compact,v}^{(t)} + \epsilon}
+
(1-\alpha) S_{drift,v}^{(t)}
\end{equation}
where $\alpha \in [0,1]$ controls the trade-off between
representation quality and update contribution.
This dynamic importance score is periodically updated
during federated training and serves as the basis for
adaptive bit allocation in subsequent communication
rounds.

\subsection{Adaptive Bit Allocation}

\textbf{Bit Allocation Rule}
The communication efficiency of the proposed algorithm 
relies on an adaptive bit allocation strategy guided by 
the dynamic importance score of each view. Instead of 
assigning a fixed quantization precision, the number of 
quantization bits is dynamically adjusted according to 
the estimated importance of the transmitted data. Given 
the importance score $S_v^{(t)}$ computed at 
communication round $t$, each view is allocated a 
quantization bit-width $b_v^{(t)}$ that balances 
clustering performance and communication cost: views with 
higher importance are assigned larger bit-widths, while 
less informative views are transmitted using lower 
precision. This adaptive mechanism enables 
communication-aware learning without sacrificing 
clustering quality. Formally, the adaptive bit allocation 
is defined as:

\begin{equation}
\label{eq:bit_allocation2}
b_v^{(t)}
=
\left\lfloor
B \cdot
\frac{s_v^{(t)}}{\sum_{j=1}^N s_j^{(t)}}
\right\rfloor
\end{equation}

\textbf{Quantization Operator}
After determining the allocated bit-width, each view 
applies a non-uniform quantization operator to compress 
its method-dependent transmission data $\mathcal{T}_v^{(t)}$ 
before transmission. The quantization process maps 
continuous values into a finite discrete representation 
defined by the selected precision. Formally, the 
transmitted data is transformed using a quantization 
operator that minimizes reconstruction distortion under 
the assigned bit constraint. This operation significantly 
reduces the transmitted payload while preserving the 
structural information required for global clustering 
convergence. The quantization process can therefore be 
mathematically written as:

\begin{equation}
\label{eq:quantization2}
\tilde{\mathcal{T}}_v^{(t)} =
Q_{b_v^{(t)}}(\mathcal{T}_v^{(t)})
\end{equation}

\subsection{View Aware Adaptive Quantized Algorithm}
\begin{algorithm}[ht]
\caption{View-Aware Adaptive Quantized FedMVC}
\label{alg:VAQ-FMC}
\begin{algorithmic}[1]

\STATE \textbf{Input:} $\{Z^v\}_{v=1}^V$, cluster number $K$, $B$, $\alpha$, $T$
\STATE \textbf{Output:}  cluster assignement $Y$
\STATE \textbf{Initialization}

\FORALL{clients $v$ in parallel}
    \STATE Solve local clustering using Eq.~(\ref{eq:local_clustering})
    \STATE Compute initial importance using Eq.~(\ref{eq:init_importance})
    \STATE Send quantized representations to server
\ENDFOR

\STATE Server initializes $C^{(0)}$ and $Y^{(0)}$

\STATE \textbf{Federated Training}

\FOR{$t = 1$ to $T$}

    \STATE Server broadcasts $C^{(t-1)}$, $Y^{(t-1)}$

    \FORALL{clients $v$ in parallel}
        \STATE Update centroids using Eq.~(\ref{eq:centroid_update})
        \STATE Compute compactness via Eq.~(\ref{eq:compactness})
        \STATE Compute drift via Eq.~(\ref{eq:drift})
        \STATE Compute importance via Eq.~(\ref{eq:importance})
        \STATE Send $s_v^{(t)}$ to server
    \ENDFOR

    \STATE Server computes bit allocation via Eq.~(\ref{eq:bit_allocation})

    \FORALL{clients $v$}
        \STATE Quantize local transmission data $\mathcal{T}_v^{(t)}$ using $b_v^{(t)}$ bits via Eq.~\eqref{eq:quantization}
        \STATE Transmit $\tilde{\mathcal{T}}_v^{(t)}$ to server
    \ENDFOR
    \STATE Server dequantizes $\{\tilde{\mathcal{T}}_v^{(t)}\}_{v=1}^V$
    \STATE Aggregate centroids
    \STATE Update assignments

\ENDFOR\\
 \RETURN $Y$
\end{algorithmic}
\end{algorithm}

Algorithm~\ref{alg:VAQ-FMC} summarizes the proposed View-Aware 
Adaptive Quantized FedMVC procedure. 
During initialization, each client solves its local clustering 
objective, computes an initial importance score, and transmits 
quantized local representations to the server, which initializes 
the global centroids $C^{(0)}$ and assignment matrix $Y^{(0)}$.

At each communication round $t$, the server broadcasts the 
current global clustering information to all clients. Each 
client updates its local centroids, computes its compactness 
loss and centroid drift, and combines them into a dynamic 
importance score $s_v^{(t)}$ (Eq.~\eqref{eq:importance}), which 
is sent to the server. The server then computes the bit budget 
$b_v^{(t)}$ allocated to each view (Eq.~\eqref{eq:bit_allocation}). 
Each client quantizes its method-dependent transmission data 
accordingly and sends it to the server, which dequantizes the 
received data, aggregates the centroids to update $C^{(t)}$, 
and updates the global assignment matrix $Y^{(t)}$. This process 
is repeated for $T$ rounds, after which the final assignment 
$Y$ is returned.

\subsection{Complexity Analysis}
At each communication round, the per-client computational 
cost is dominated by the centroid update (Eq.~\eqref{eq:centroid_update}) 
and the compactness loss (Eq.~\eqref{eq:compactness}), both 
involving matrix operations of size $n \times K \times r$, 
resulting in a complexity of $O(nKr)$. The centroid drift 
(Eq.~\eqref{eq:drift}) only involves a comparison between two 
$K \times r$ matrices, with negligible complexity $O(Kr)$. 
The importance score and bit allocation (Eqs.~\eqref{eq:importance}, 
\eqref{eq:bit_allocation}) involve only scalar operations across 
$V$ views, with complexity $O(V)$. Overall, the per-round 
computational complexity of the proposed algorithm is 
$O(nKr)$, which matches the complexity of the underlying local 
clustering step and therefore introduces negligible computational 
overhead compared to the base FedMVC method.

In terms of communication, the proposed VAQ 
does not alter the number of transmitted elements, but only 
the number of bits $b_v^{(t)}$ used to represent each element 
of $\mathcal{T}_v^{(t)}$. For most FedMVC methods, 
$|\mathcal{T}_v^{(t)}| = |Z^v| = n \times r$. The per-round 
communication cost for client $v$ is therefore 
$b_v^{(t)} \cdot |\mathcal{T}_v^{(t)}|$. While uniform 
quantization fixes $b_v^{(t)} = 8$ for all views, leading to 
a total budget $B_{\text{uniform}} = 8 \times V$, the proposed 
approach allows the total budget $B$ to be set independently, 
typically satisfying $B \leq B_{\text{uniform}}$, while 
redistributing it across views according to their importance 
scores. This flexibility enables the proposed approach to 
achieve a lower overall communication cost than uniform 
quantization, while preserving and in some cases improving clustering performance by allocating more bits to the most 
informative views.

\section{EXPERIMENTS}
\label{sec:results}
\subsection{Experiment setting}
\textbf{Datasets.} We conduct experiments on five multi-view datasets with diverse 
characteristics widely used in the literature:
\textbf{Scene}\cite{fei2005bayesian} is a multi-view benchmark dataset for scene recognition, where 
heterogeneous visual features capturing complementary aspects of scene 
characteristics are treated as distinct views, containing 4485 samples 
distributed across 3 views and 15 semantic scene categories. 
\textbf{MNIST-USPS}\cite{peng2019comic} is a cross-domain handwritten digit dataset comprising 5000 samples with 2 views and 10 clusters. 
\textbf{BDGP}\cite{cai2012joint} contains 2500 samples, 2 views, and 5 clusters.
\textbf{HW}\footnote{\url{https://archive.ics.uci.edu/ml/datasets/Multiple+Features/}} (Handwritten)\cite{van1998handwritten} consists of 2000 samples described by 6 views and organized into 10 clusters. 
Finally, \textbf{Caltech-5V}\cite{fei2004learning} includes 1400 samples represented by 5 views across 7 clusters. 
The statistics of these datasets are summarized in Table \ref{tab:datasets}.
\begin{table}[ht]
\centering
\caption{Statistics of the benchmark multi-view clustering datasets used in our experiments, including the number of samples, views, and clusters.}
\label{tab:datasets}
\begin{tabular}{|c|c|c|c|}
\hline
Dataset & Samples & Views & Clusters \\
\hline
MNIST-USPS & 5000 & 2 & 10 \\
\hline
Scene & 4485 & 3 & 15 \\
\hline
BDGP  & 2500 & 2 & 5 \\
\hline
HW & 2000 & 6 & 10 \\
\hline
Caltech-5V & 1400 & 5 & 7 \\
\hline
\end{tabular}
\end{table}

\textbf{Comparison Methods.} To evaluate the generality and effectiveness of our VAQ 
framework, we integrate it into several state-of-the-art federated 
multi-view clustering methods: ESFMC \cite{chen2025effective}, FCUIF \cite{ren2024novel}, FMCSC \cite{chen2024bridging}, and 
HFMVC \cite{jiang2024heterogeneity}. These methods represent diverse approaches to multi-view 
clustering in federated settings.
To enable consistent and fair comparison across all methods, we 
develop modular quantization components that can be seamlessly integrated into 
any FedMVC framework. Specifically, we implement two 
quantization modules: (1) a uniform quantization module that applies standard 
8-bit integer quantization to all shared representations, and (2) our proposed 
view-aware adaptive quantization module that dynamically allocates bits based on 
view-specific importance metrics.\\
\textbf{Evaluation Scenarios.} For each method-dataset combination, we evaluate three 
scenarios: (1) Baseline: no quantization (float32 representations), (2) Uniform 
Quantization: fixed 8-bit quantization for all views, and (3) View-Aware 
Quantization: adaptive bit allocation with a total budget distributed 
across views. 
To evaluate the benefit of view-aware bit allocation 
independently of the total budget, we consider two 
complementary settings. In the first setting, the total 
bit budget is fixed to $B = V \times 8$ bits, matching 
the budget of uniform int8 quantization, so that any 
observed difference in performance is solely attributable 
to the allocation strategy. In the second setting, $B$ is set to values lower than 
or equal to $V \times 8$ bits (i.e., $B \leq B_{\text{uniform}}$), 
allowing the proposed approach to achieve a lower or equal 
communication cost than uniform quantization while maintaining 
competitive clustering performance.
as well as, all methods are evaluated using the same 
federated data partition and identical view distribution across clients. 
Communication overhead is precisely tracked using our custom communication 
tracking module that accounts for all data transmitted between clients and server.\\
\textbf{Evaluation Metrics.} To evaluate clustering performance, we adopt three standard metrics: accuracy (ACC), normalized mutual information (NMI), and adjusted rand index (ARI), with higher values reflecting superior clustering results.
\subsection{Clustering Results}
In this section, we present and analyze the experimental 
results of the proposed VAQ 
algorithm. We first evaluate the clustering performance 
of VAQ compared to the baseline (no quantization) and 
uniform int8 quantization across four FedMVC methods and 
five benchmark datasets. We then analyze the communication 
overhead achieved by each scenario and examine the 
communication-performance trade-off under different bit 
budgets.

\begin{table*}[t]
\centering
\caption{Clustering Performance Comparison Under Different 
Compression Scenarios Across Four FedMVC Methods and Five 
Benchmark Datasets. \textbf{Bold} results indicate that 
View-Aware-Q achieves equal or superior performance 
compared to the baseline, while \underline{underlined} 
values denote a marginal performance reduction. 
$^\dagger$For View-Aware-Q, the total bit budget satisfies 
$B \leq V \times 8$ bits.}
\label{tab:clustering_performance}
\renewcommand{\arraystretch}{1.2}
\resizebox{\textwidth}{!}{%
\begin{tabular}{llccccccccccccccc}
\toprule
\multirow{2}{*}{\textbf{Method}} & 
\multirow{2}{*}{\textbf{Scenario}} & 
\multicolumn{3}{c}{\textbf{BDGP}} & 
\multicolumn{3}{c}{\textbf{HW}} & 
\multicolumn{3}{c}{\textbf{MNIST-USPS}} & 
\multicolumn{3}{c}{\textbf{Scene-15}} & 
\multicolumn{3}{c}{\textbf{Caltech-5V}} \\
\cmidrule(lr){3-5} \cmidrule(lr){6-8} 
\cmidrule(lr){9-11} \cmidrule(lr){12-14} 
\cmidrule(lr){15-17}
& & ACC & NMI & ARI & ACC & NMI & ARI & 
ACC & NMI & ARI & ACC & NMI & ARI & 
ACC & NMI & ARI \\
\midrule
\multirow{3}{*}{FMCSC} 
& None      & 92.88 & 80.88 & 83.42 & 93.60 & 86.80 & 86.35 & 94.06 & 88.34 & 87.53 & 39.49 & 37.44 & 21.61 & 82.57 & 70.33 & 67.76 \\
& Uniform-Q & 94.24 & 83.51 & 86.38 & 92.45 & 85.53 & 84.12 & 92.88 & 87.51 & 85.50 & 39.40 & 37.62 & 21.52 & 82.00 & 70.73 & 66.93 \\
& View-Aware-Q$^\dagger$ & \textbf {94.60}
& \textbf {84.58}
& \textbf {87.06}
& \textbf {94.30} 
& \textbf {88.17}
& \textbf {87.76}
& \textbf {94.44}
& \textbf {88.82}
& \textbf {88.26}
& \textbf {40.65}
& \textbf {37.24}
& \textbf {21.74}
& \textbf {84.43}
& \textbf {72.39}
& \textbf {69.88} \\
\midrule
\multirow{3}{*}{FCUIF} 
& None      & 66.80 & 38.51 & 35.72 & 63.35 & 58.97 & 47.77 & 50.66 & 47.27 & 34.53 & 30.58 & 30.94 & 17.62 & 46.43 & 40.22 & 30.01 \\
& Uniform-Q & 52.32 & 30.49 & 25.56 & 62.85 & 64.66 & 53.42 & 51.30 & 49.05 & 35.76 & 32.34 & 31.03 & 18.71 & 54.50 & 42.04 & 29.35 \\
& View-Aware-Q$^\dagger$ & \textbf {81.24}
& \textbf {59.49}
& \textbf {60.16}
& \textbf {65.55}
& \textbf {62.25}
& \textbf {53.76}
& \textbf {54.04} 
& \textbf {48.55}
& \textbf {37.06}
& \textbf {34.99}
& \textbf {31.83}
& \textbf {19.16}
& \textbf {51.21}
& \textbf {40.44}
& \textbf {30.95} \\
\midrule
\multirow{3}{*}{HFMVC} 
& None      & 94.16 & 83.46 & 85.83 & 95.20 & 90.29 & 89.73 & 75.94 & 77.14 & 66.61 & 33.36 & 35.04 & 20.40 & 58.93 & 53.53 & 42.45 \\
& Uniform-Q & 94.32 & 83.84 & 86.18 & 94.70 & 89.79 & 88.78 & 76.42 & 77.44 & 67.06 & 33.53 & 35.66 & 21.15 & 58.93 & 53.57 & 42.47 \\
& View-Aware-Q$^\dagger$ & \underline {93.96}
& \underline {82.98}
& \underline {85.38}
& \underline {95.05}
& \underline {90.16}
& \underline {89.45}
& \textbf {76.12}
& \textbf {77.49}
& \textbf {67.05}
& \textbf {33.78}
& \textbf {36.08}
& \textbf {21.41}
& \textbf {71.93}
& \textbf {56.51}
& \textbf {48.89} \\
\midrule
\multirow{3}{*}{ESFMC} 
& None      & 98.28 & 94.04 & 95.76 & 97.50 & 94.42 & 94.54 & 94.52 & 88.59 & 88.33 & 31.68 & 36.50 & 18.27 & 90.86 & 84.43 & 82.63 \\
& Uniform-Q & 97.04 & 91.91 & 92.86 & 97.55 & 94.69 & 94.64 & 93.58 & 86.91 & 86.48 & 34.09 & 35.42 & 19.10 & 88.71 & 81.59 & 79.47 \\
& View-Aware-Q$^\dagger$ & \underline {98.04}
& \underline {93.79}
& \underline {95.18}
& \underline {97.20} 
& \underline {93.85} 
& \underline {93.89}
& 91.02
& 85.40
& 82.89
& \textbf {34.92}
& \underline {36.14}
& \textbf {19.93}
& \textbf {90.86}
& \underline {83.49}
& \underline {82.27} \\
\bottomrule
\multicolumn{17}{l}{$^\dagger$ $B \leq V \times 8$ bits.}
\end{tabular}%
}
\end{table*}

\subsubsection{Main Clustering Results}
Table~\ref{tab:clustering_performance} reports the clustering 
performance of the three compression scenarios: baseline 
(no quantization), uniform int8 quantization (Uniform-Q), 
and the proposed VAQ across four FedMVC methods and five benchmark datasets, 
evaluated in terms of ACC, NMI, and ARI.

Overall, the results demonstrate that View-Aware-Q 
consistently maintains or improves clustering performance 
compared to the baseline across most methods and datasets, 
while operating under a reduced communication budget 
($B \leq V \times 8$ bits). In several cases, View-Aware-Q 
even surpasses the baseline, notably for FCUIF on BDGP 
(ACC: 81.24\% vs. 66.80\%) and FMCSC on BDGP 
(ACC: 94.60\% vs. 92.88\%), suggesting that the adaptive 
bit allocation acts as an implicit regularization mechanism 
that can improve clustering quality.

These results confirm that allocating bits proportionally 
to view importance is more effective than uniform 
allocation, even under a tighter communication budget. 
For instance, View-Aware-Q achieves an ACC gain of 
$+28.92\%$ over Uniform-Q for FCUIF on BDGP 
(ACC: 81.24\% vs. 52.32\%), demonstrating that adaptive 
bit allocation better preserves the discriminative 
information of informative views compared to uniform 
compression.

In Table~\ref{tab:clustering_performance}, results in 
\textbf{bold} indicate that View-Aware-Q achieves equal 
or superior performance compared to the baseline, while 
\underline{underlined} values denote a marginal performance 
reduction. As can be observed, the vast majority of 
results are either bolded or underlined, confirming that 
View-Aware-Q preserves clustering quality across nearly 
all method-dataset combinations under a reduced 
communication budget.

The only notable exception is observed for ESFMC on 
MNIST-USPS, where View-Aware-Q yields a slight 
performance drop compared to the baseline. This behavior 
appears to be method-specific, as the same compression 
settings consistently improve performance for all other 
methods on this dataset, suggesting that this sensitivity 
is related to the internal dynamics of ESFMC rather than 
a limitation of the proposed quantization strategy.

\subsubsection{Communication Performance Trade-off}
Having established that View-Aware-Q maintains or improves 
clustering performance compared to the baseline 
(Table~\ref{tab:clustering_performance}), we now examine 
the associated communication overhead in 
Table~\ref{tab:comm_overhead}. The results reveal that 
this performance is achieved at a significantly reduced 
communication cost. Across all methods and datasets, 
View-Aware-Q consistently achieves higher communication 
reduction than Uniform-Q, with a maximum reduction ratio 
of \textbf{6.59×} for HFMVC on Caltech-5V. These 
results confirm that the proposed view-aware bit 
allocation effectively reduces communication overhead 
while maintaining and in several cases surpassing baseline clustering performance, demonstrating a 
favorable communication-performance trade-off.

It is worth noting that for some methods, the pretrain 
cost per round differs from the per-round cost during 
federated training, which impacts the total communication 
cost. In only three cases: ESFMC on Scene-15, FCUIF on 
MNIST-USPS, and FMCSC on MNIST-USPS. View-Aware-Q 
achieves a reduction ratio equal or approximately equal to 
Uniform-Q. This is explained by the fact that the total 
bit budget for View-Aware-Q in these configurations 
equals $B = V \times 8$ bits, identical to that of 
Uniform-Q, so that the communication reduction is 
solely determined by the quantization scheme rather 
than a reduced budget.

\begin{table*}[t]
\centering
\caption{Communication Overhead Comparison Under Different 
Compression Scenarios Across Four FedMVC Methods and Five 
Benchmark Datasets. Round: per-round communication cost (MB); 
Total: total communication cost (MB); Red.: reduction ratio 
vs. baseline. \textbf{Bold} values indicate that View-Aware-Q 
achieves a strictly higher reduction ratio than Uniform-Q, 
while \underline{underlined} values denote a negligible 
difference. $^\dagger$The total bit budget for View-Aware-Q 
corresponds to the same configuration as in 
Table~\ref{tab:clustering_performance}.}
\label{tab:comm_overhead}
\renewcommand{\arraystretch}{1.2}
\resizebox{\textwidth}{!}{%
\begin{tabular}{llccccccccccccccc}
\toprule
\multirow{2}{*}{\textbf{Method}} & 
\multirow{2}{*}{\textbf{Scenario}} & 
\multicolumn{3}{c}{\textbf{BDGP}} & 
\multicolumn{3}{c}{\textbf{HW}} & 
\multicolumn{3}{c}{\textbf{MNIST-USPS}} & 
\multicolumn{3}{c}{\textbf{Scene-15}} & 
\multicolumn{3}{c}{\textbf{Caltech-5V}} \\
\cmidrule(lr){3-5} \cmidrule(lr){6-8} 
\cmidrule(lr){9-11} \cmidrule(lr){12-14} 
\cmidrule(lr){15-17}
& & Round & Total & Red. & 
Round & Total & Red. & 
Round & Total & Red. & 
Round & Total & Red. & 
Round & Total & Red. \\
\midrule
\multirow{3}{*}{FMCSC} 
& None      & 113.48 & 624.12 & — & 799.1003 & 4395.0518 & — & 109.49 & 602.20 & — & 195.13 & 1073.23 & — & 676.28 & 3719.57 & — \\
& Uniform-Q & 28.37 & 156.03 & 4.00× & 199.78 & 1098.76 & 4.00× & 27.37 & 150.55 & 4.00× & 48.78 & 268.31 & 4.00× & 169.07 & 929.89 & 4.00× \\
& View-Aware-Q$^\dagger$ & 27.03 & 149.35 & \textbf{4.18×} & 191.59 & 1057.82 & \textbf{4.15×} & 27.47 & 151.02 & \underline{3.99×} & 47.71 & 262.95 & \textbf{4.08×} & 164.83 & 908.69 & \textbf{4.09×} \\
\midrule
\multirow{3}{*}{FCUIF} 
& None      & 0.1931 & 1.7405 & — & 0.4635 & 3.2501 & — & 0.3862 & 3.4809 & — & 0.5196 & 2.0847 & — & 0.2704 & 1.8959 & — \\
& Uniform-Q & 0.0501 & 0.4530 & 3.84× & 0.1202 & 0.8469 & 3.84× & 0.1001 & 0.9060 & 3.84× & 0.1347 & 0.5452 & 3.83× & 0.0701 & 0.4940 & 3.84× \\
& View-Aware-Q$^\dagger$ & 0.0441 & 0.4053 & \textbf{4.29×} & 0.1098 & 0.7849 & \textbf{4.14×} & 0.1001 & 0.9060 & \underline{3.84×} & 0.1240 & 0.5132 & \textbf{4.06×} & 0.0498 & 0.3722 & \textbf{5.09×} \\
\midrule
\multirow{3}{*}{HFMVC} 
& None      & 2.4462 & 22.0156 & — & 5.8708 & 41.0957 & — & 4.8923 & 83.1699 & — & 6.5827 & 59.2439 & — & 3.4246 & 30.8218 & — \\
& Uniform-Q & 0.6151 & 5.5361 & 3.98× & 1.4763 & 10.3340 & 3.98× & 1.2302 & 20.9141 & 3.98× & 1.6553 & 14.8976 & 3.98× & 0.8612 & 7.7505 & 3.98× \\
& View-Aware-Q$^\dagger$ & 0.4959 & 4.4298 & \textbf{4.97×} & 0.9880 & 6.9160 & \textbf{5.94×} & 0.9251 & 15.7261 & \textbf{5.29×} & 1.2447 & 11.2021 & \textbf{5.29×} & 0.5194 & 4.6743 & \textbf{6.59×} \\
\midrule
\multirow{3}{*}{ESFMC} 
& None      & 0.6151 & 4.9210 & — & 1.4763 & 8.8577 & — & 1.2302 & 19.6838 & — & 1.6553 & 6.6211 & — & 0.8612 & 13.7787 & — \\
& Uniform-Q & 0.1574 & 1.2589 & 3.91× & 0.3777 & 2.2659 & 3.91× & 0.3147 & 5.0354 & 3.91× & 0.4234 & 2.1172 & 3.13× & 0.2203 & 3.5248 & 3.91× \\
& View-Aware-Q$^\dagger$ & 0.1192 & 0.9537 & \textbf{5.16×} & 0.2861 & 1.7166 & \textbf{5.16×} & 0.2766 & 4.4250 & \textbf{4.45×} & 0.4234 & 2.1172 & \underline{3.13×} & 0.1348 & 2.1576 & \textbf{6.39×} \\
\bottomrule
\end{tabular}%
}
\end{table*}
\subsubsection{Analysis and Discussion}
\begin{figure*}[t]
    \centering
    \includegraphics[width=1.5\columnwidth]{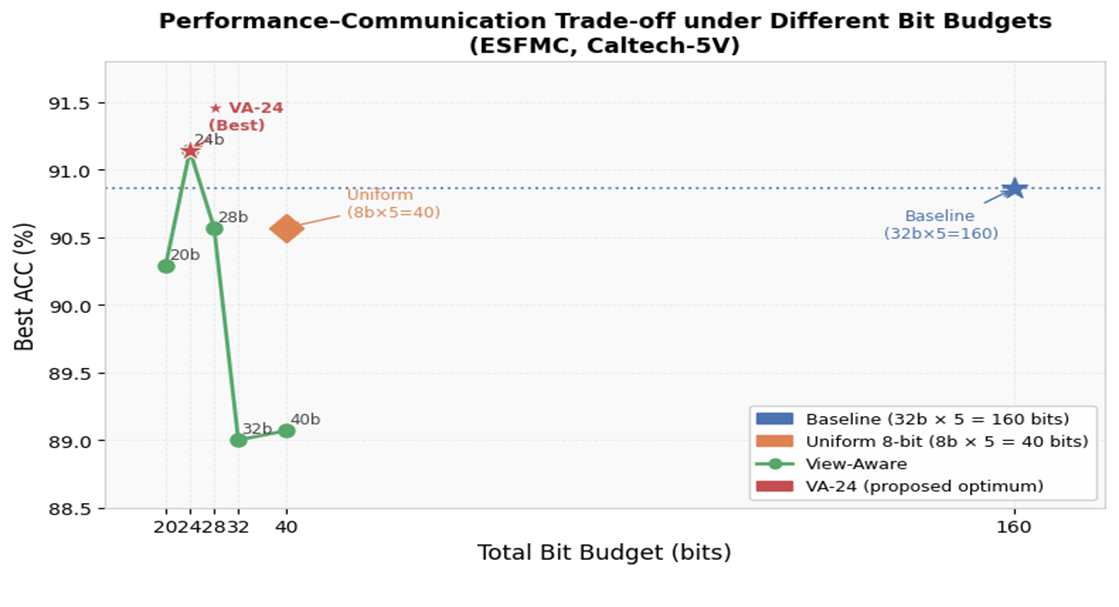}
    \caption{Best clustering accuracy (ACC) as a function 
of the total bit budget for the proposed view-aware 
quantization, compared to the baseline (no quantization) 
and uniform 8-bit quantization, on the Caltech-5V 
dataset using ESFMC. View-Aware-Q with $B=24$ bits 
achieves the best ACC (91.14\%), surpassing both the 
baseline (90.86\%) and Uniform-Q (90.57\%), while 
reducing the communication cost by 6.39$\times$ compared 
to the baseline.}
    \label{fig:tradeoff}
\end{figure*}

To illustrate how the total bit budget $B$ influences 
the clustering performance, Fig.~\ref{fig:tradeoff} 
shows the best ACC as a function of $B$ for Method1 
on Caltech-5V. As can be observed, reducing $B$ below 
the uniform budget ($V \times 8 = 40$ bits) does not 
necessarily lead to performance degradation. In 
particular, View-Aware-Q with $B = 24$ bits achieves 
an ACC of 91.14\%, surpassing both the baseline 
(90.86\%) and Uniform-Q (90.57\%), while reducing the 
communication cost by \textbf{6.39×} compared to the 
baseline. However, further reducing the budget beyond 
this point (e.g., $B = 20$ bits) leads to a noticeable 
performance drop, suggesting the existence of an optimal 
bit budget that balances communication efficiency and 
clustering quality. This figure serves as an illustrative 
example of how the total bit budget is selected for each 
method-dataset combination in our experiments.

A key strength of the proposed View-Aware-Q algorithm 
is its ability to seamlessly integrate into existing 
FedMVC methods without modifying their internal 
architecture. To validate this generality, we evaluated 
View-Aware-Q on four FedMVC methods with heterogeneous 
architectures, differing in their encoder design, 
transmitted data, and aggregation strategy. As shown 
in Tables~\ref{tab:clustering_performance} and 
\ref{tab:comm_overhead}, View-Aware-Q consistently 
maintains or improves clustering performance while 
reducing communication overhead across all four methods 
and five benchmark datasets. This confirms that the 
proposed bit allocation strategy is agnostic to the 
underlying FedMVC method, making it a flexible and 
practical solution for communication-efficient federated 
multi-view clustering.

\begin{figure*}[]
    \centering
    
    \begin{subfigure}[b]{0.48\textwidth}
        \centering
        \caption{View Importance Score evolution across communication 
                 rounds for the HW dataset (6 views) under the FMCSC 
                 FedMVC framework. The dashed 
                 line represents the uniform reference score .}
        \includegraphics[width=\textwidth]{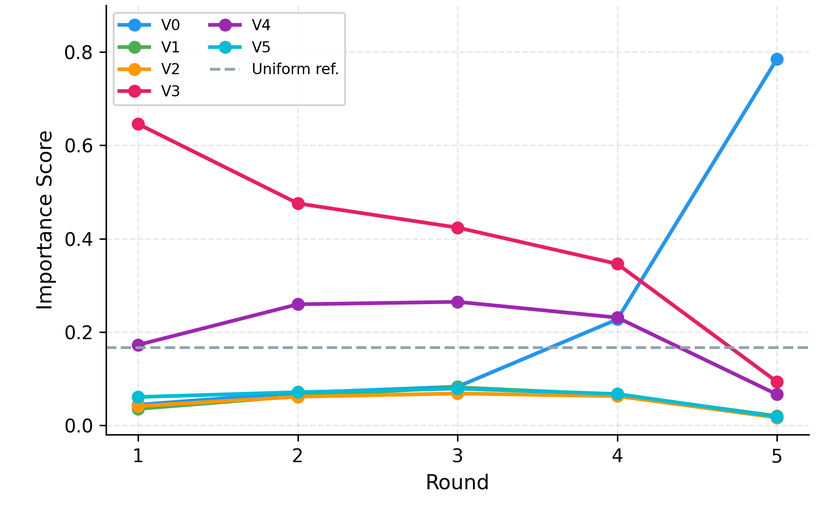}
        \label{fig:view_importance_score}
    \end{subfigure}
    \hfill
    \begin{subfigure}[b]{0.48\textwidth}
        \centering
        \caption{Per-round bit allocation per view for the HW dataset 
                 under FMCSC with VAQ. The dashed 
                 line represents the uniform int8 baseline (8 bits per 
                 view). Bit allocations satisfy 
                 $\sum_{v=1}^{V} b_v \leq B_{total} \leq V \times 8 = 48$ 
                 bits.}
        \includegraphics[width=\textwidth]{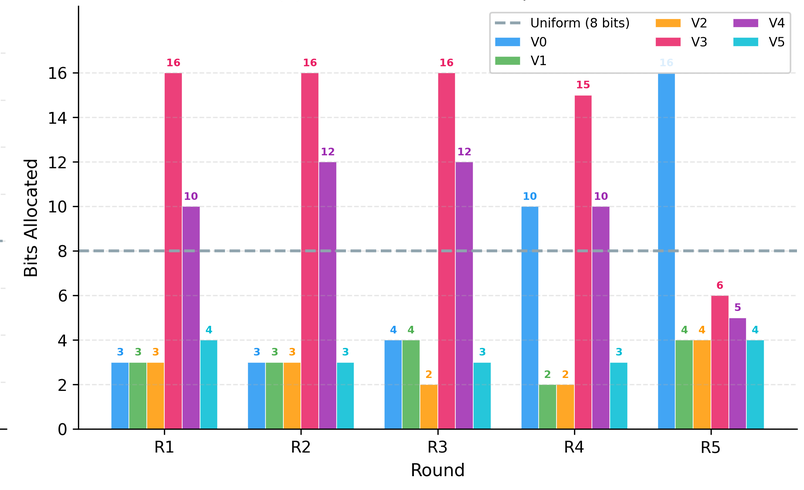}
        \label{fig:bits_allocated}
    \end{subfigure}

    \caption{VAQ analysis in FMCSC on the HW dataset
             ($V=6$, $B_{total} \leq V \times 8 = 48$ bits). At each 
             federated round, FMCSC estimates the importance of each 
             client view and adaptively redistributes the bit budget 
             accordingly, in contrast to the fixed uniform int8 
             allocation.}
    \label{fig:view_importance}
\end{figure*}

\paragraph{View Importance and Bit Allocation Dynamics.}
Figure~\ref{fig:view_importance_score} illustrates the evolution of view 
importance scores across communication rounds for the HW dataset (6 views). 
In the early rounds (R1--R3), view $\mathcal{V}_3$ dominates with an 
importance score exceeding 0.42, reflecting its tightly clustered 
representations (lowest compactness loss $\approx$ 0.27--0.67) and moderate 
centroid drift. This dominance is directly reflected in 
Figure~\ref{fig:bits_allocated}, where $\mathcal{V}_3$ is consistently 
allocated the maximum budget of 16 bits across the first three rounds. 
Meanwhile, views $\mathcal{V}_1$, $\mathcal{V}_2$, and $\mathcal{V}_5$, 
which exhibit higher compactness losses and lower drift, receive as few as 
2--3 bits, well below the uniform baseline of 8 bits. A notable shift occurs 
at Round 4, where $\mathcal{V}_0$ begins to gain importance (0.23) as its 
compactness loss drops sharply from 8.19 to 1.07, indicating that its local 
cluster structure is rapidly improving --- a dynamic that a static uniform 
allocation would fail to capture. By Round 5, $\mathcal{V}_0$ reaches the 
highest importance score (0.79) and is allocated the full 16-bit budget, 
while $\mathcal{V}_3$ drops to only 6 bits, reflecting a role reversal 
driven by the convergence dynamics of the federated training. This adaptive 
behavior demonstrates that VAQ successfully tracks the 
evolving informativeness of each view across rounds, redistributing the bit 
budget where precision matters most, rather than applying a fixed allocation 
as in uniform quantization. It is worth noting that $B_{total} \leq V \times 8$ 
serves as an upper bound on the total bits allocated; the actual sum satisfies 
$\sum_{v=1}^{V} b_v \leq B_{total}$ due to integer rounding during 
proportional allocation and a per-view cap of 16 bits, which prevents any 
single view from monopolizing the communication budget. This guarantee ensures 
that the communication overhead of VAQ never exceeds that 
of uniform int8 quantization, while achieving a more informed distribution of 
precision across views.

\subsection{Ablation Study}
\textbf{Component-wise Ablation.}
Table~\ref{tab:ablation} reports the ablation study 
results on Method3 evaluated on the HW dataset. Due 
to computational constraints, this dataset-method 
combination is used as a representative case. The 
results demonstrate a clear and consistent improvement 
as each component is added. Compared to VAQ-Uniform, 
VAQ-Static achieves a gain of $+0.60\%$ in ACC by 
introducing view-aware bit allocation based on fixed 
initial importance scores, confirming that adaptive 
allocation is beneficial even without dynamic updates. 
Furthermore, VAQ-Full outperforms VAQ-Static by an 
additional $+1.25\%$ in ACC by incorporating dynamically 
updated importance scores, demonstrating the value of 
iteratively refining the bit allocation throughout 
training. Notably, VAQ-Full also achieves the lowest 
communication cost (1057.82 MB vs. 1098.76 MB for 
VAQ-Uniform), confirming that the proposed dynamic 
view-aware allocation simultaneously improves clustering 
performance and reduces communication overhead. These 
results validate the contribution of each component 
of the proposed VAQ algorithm.

\begin{table}[ht]
\centering
\caption{Ablation Study on the Contribution of Each 
Component of the Proposed VAQ Algorithm (FMCSC, HW 
Dataset). \textbf{Bold} values indicate the best result.}
\label{tab:ablation}
\renewcommand{\arraystretch}{1.2}
\resizebox{\columnwidth}{!}{%
\begin{tabular}{lcccccc}
\toprule
\textbf{Variant} & \textbf{ACC (\%)} & \textbf{NMI (\%)} & 
\textbf{ARI (\%)} & \textbf{Total Cost (MB)} & 
\textbf{Red.} \\
\midrule
VAQ-Uniform & 92.45 & 85.53 & 84.12 & 1098.76 & 4.00× \\
VAQ-Static  & 93.05 & 86.21 & 85.27 & 1060.48 & 4.14× \\
VAQ-Full    & \textbf{94.30} & \textbf{88.17} & 
\textbf{87.76} & \textbf{1057.82} & \textbf{4.15×} \\
\bottomrule
\multicolumn{6}{l}{\small VAQ-Uniform: uniform bit 
allocation ($b_v = B/V$).}\\
\multicolumn{6}{l}{\small VAQ-Static: view-aware 
allocation with fixed initial importance scores.}\\
\multicolumn{6}{l}{\small VAQ-Full: proposed algorithm 
with dynamically updated importance scores.}\\
\end{tabular}
}
\end{table}

\textbf{Sensitivity to the Weighting Factor $\alpha$.}
Table~\ref{tab:alpha_sensitivity} reports the sensitivity 
of the proposed VAQ algorithm to the weighting factor 
$\alpha$, which controls the balance between clustering 
compactness and centroid drift in the importance score 
(Eq.~\ref{eq:importance}). Specifically, $\alpha = 0$ 
relies solely on centroid drift, $\alpha = 1$ relies 
solely on clustering compactness, and intermediate 
values combine both criteria. The results show that 
$\alpha = 0.5$ achieves the best clustering performance 
across all metrics (ACC: 94.65\%, NMI: 88.59\%, ARI: 
88.51\%), confirming that both compactness and centroid 
drift contribute complementary information to the 
importance score. Relying solely on centroid drift 
($\alpha = 0$) or solely on compactness ($\alpha = 1$) 
leads to suboptimal performance, with ACC dropping to 
93.80\% and 91.14\% respectively. These results validate 
the design choice of combining both criteria in the 
importance score, and suggest that $\alpha = 0.5$ 
provides a robust and balanced configuration. 
Accordingly, $\alpha = 0.5$ is used in all experiments 
reported in this paper.

\begin{table}[t]
\centering
\caption{Sensitivity Analysis of the Weighting Factor 
$\alpha$ on FMCSC evaluated on the HW Dataset. \textbf{Bold} 
values indicate the best result.}
\label{tab:alpha_sensitivity}
\renewcommand{\arraystretch}{1.2}
\resizebox{\columnwidth}{!}{%
\begin{tabular}{lccccc}
\toprule
\textbf{$\alpha$} & \textbf{ACC (\%)} & \textbf{NMI (\%)} & 
\textbf{ARI (\%)} & \textbf{Total Cost (MB)} \\
\midrule
0.0  & 93.80 & 87.21 & 86.79 & 1063.71 \\
0.25 & 92.05 & 84.57 & 83.40 & 1057.10 \\
0.5  & \textbf{94.65} & \textbf{88.59} & 
      \textbf{88.51} & 1058.15 \\
0.75 & 92.10 & 85.55 & 83.55 & 1066.88 \\
1.0  & 91.14 & 83.48 & 81.66 & 1071.59 \\
\bottomrule
\end{tabular}%
}
\end{table}

\section{Conclusion}
\label{sec:conclusion}
In this paper, we addressed the problem of communication 
overhead in FedMVC by proposing 
a VAQ algorithm that adaptively 
allocates quantization bits across views according to 
their dynamically estimated importance. The importance 
score combines two complementary criteria: clustering 
compactness and centroid drift, to capture both the 
static quality and the dynamic behavior of each view 
throughout the federated training process. The proposed 
algorithm is formulated as a constrained optimization 
problem under a global bit budget constraint, and can 
be seamlessly integrated into any existing FedMVC method 
without modifying its internal architecture.

In future work, we plan to automate the selection of 
the total bit budget by formulating it as a constrained 
optimization problem, eliminating the need for manual 
tuning across method-dataset combinations.

Ablation studies confirm the contribution of each 
component of the proposed algorithm, and sensitivity 
analysis validates the choice of $\alpha = 0.5$ as 
a robust configuration. Extensive experiments conducted 
on five benchmark datasets across four state-of-the-art 
FedMVC methods demonstrate that VAQ achieves up to 
6.59× communication reduction compared to the baseline, 
while maintaining or surpassing baseline clustering 
performance in the vast majority of cases.

\section*{Declaration of generative AI and AI-assisted technologies in the manuscript preparation process. }

During the preparation of this work, the authors used 
Claude in order to facilitate the search 
and screening of the scientific literature, improve 
the language and readability of the manuscript, and 
correct grammar. After using this tool, the authors 
reviewed and edited the content as needed, verified 
the accuracy of all references and information, and 
take full responsibility for the content of the 
publication.

\clearpage 

\section{}\label{}

\printcredits




\end{document}